%% file: main.tex
\documentclass[10pt,twocolumn,letterpaper]{article}

\usepackage{iccv}

\usepackage[utf8]{inputenc} 
\usepackage[T1]{fontenc}    
\usepackage[pagebackref=true,breaklinks=true,letterpaper=true,colorlinks,bookmarks=false]{hyperref}      
\usepackage{url}            
\usepackage{booktabs}       
\usepackage{amsfonts}       
\usepackage{nicefrac}       
\usepackage{microtype}      

\usepackage{times}
\usepackage{epsfig}
\usepackage{graphicx}
\usepackage{amsmath}
\usepackage{amssymb}
\usepackage{lipsum} 
\usepackage{color}
\usepackage{xcolor}

\usepackage{multirow}

\usepackage{xspace}

\iccvfinalcopy 

\def\iccvPaperID{4758} 

\ificcvfinal\fi

\newcommand{\method}{BAR-CNN\xspace}
\newcommand{\VCOCO}{\emph{V-COCO}\xspace}
\newcommand{\VRD}{\emph{Visual Relationships}\xspace}
\newcommand{\OID}{\emph{Open Images}\xspace}
\newcommand{\p}{p}

\begin{document}

\title{Detecting Visual Relationships Using Box Attention}

\author{
  Alexander~Kolesnikov$^1$\thanks{Work partially done at IST Austria.} \\
  Google~Research \\
  \and
  Alina~Kuznetsova$^1$ \\
  Google~Research \\
  \and
  Christoph H. Lampert$^2$ \\
  IST Austria \\
  \and
  Vittorio~Ferrari$^1$ \\
  Google~Research \\
  \and
  $^1$\texttt{\{akolesnikov,akuznetsa,vittoferrari\}@google.com} \quad $^2$\texttt{chl@ist.ac.at}
}

\maketitle

\begin{abstract}

We propose a new model for detecting visual relationships, such as "person riding motorcycle" or "bottle on table".
This task is an important step towards comprehensive structured image understanding, going beyond detecting individual objects.
Our main novelty is a \emph{Box Attention} mechanism that allows to model pairwise interactions between objects using standard object detection pipelines. 
The resulting model is conceptually clean, expressive and relies on well-justified training and prediction procedures.
Moreover, unlike previously proposed approaches, our model does not introduce any additional complex components or hyperparameters on top of those already required by the underlying detection model.
We conduct an experimental evaluation on three challenging datasets, \VCOCO, \VRD and \OID,
demonstrating strong quantitative and qualitative results.

\end{abstract}

\input{intro}

\input{rw}
\input{method}
\input{experiments}

\input{conclusion}

{
\bibliographystyle{ieee}
\bibliography{iccv}
}

\end{document}

%% file: intro.tex
\section{Introduction}

\begin{figure}
    \centering
    \includegraphics[width=0.45\textwidth]{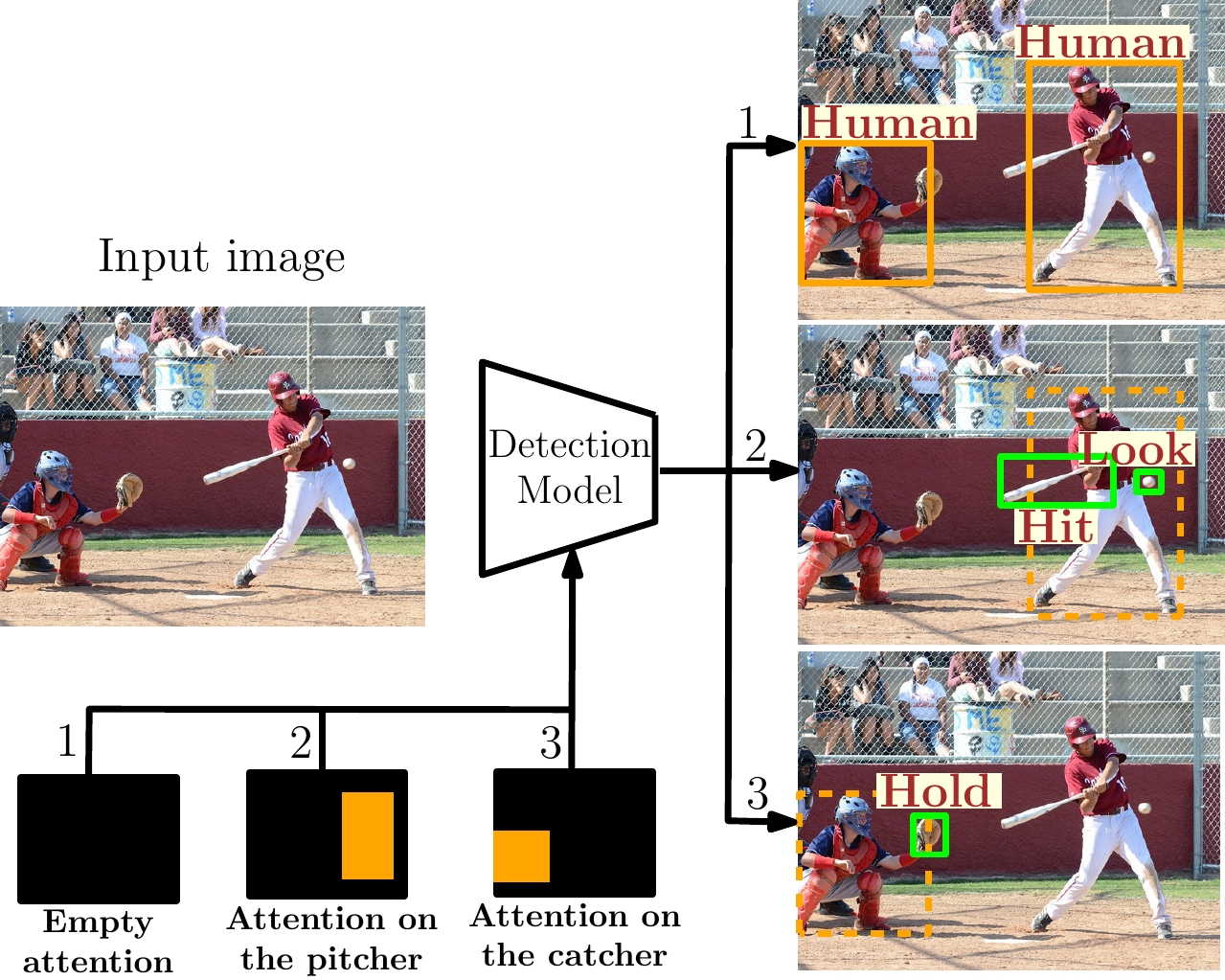}
    \caption{Schematic illustration of the proposed model.
             It is based on the general object detection pipeline augmented with our box attention mechanism.
             If an input box attention is empty (marked as $1$ in the figure), then our model predicts detections for all subjects (humans in this case).
             However, if an input box attention represents a subject's bounding box (marked as $2$ or $3$),
             then our model predicts detections for objects that interact with this subject.
             Our approach is conceptually clean, easy-to-implement and, at the same time, achieves strong performance on a wide range of visual relationship detection benchmarks.}
    \label{fig:method}
\end{figure}

The task of detecting visual relationships aims at localizing all pairs
of interacting objects in an input image and identifying relationships
between them.
The ability to recognize visual relationships is crucial for achieving
comprehensive understanding of visual scenes.  
Humans can easily analyze visual scenes and infer various relationships
between objects, such as their spatial relations (\eg \emph{bottle on table}) or
human-centric interactions (\eg \emph{person holding a cup}).
As a consequence, 
the task of detecting visual relationships has recently attracted a lot of attention
in the computer vision community~\cite{dai2017detecting,gkioxari2017detecting,gupta2015visual,
                                       krishna2018referring,li2017vip,
                                       liang2017deep,lu2016visual,peyre2017weakly,
                                       yu2017visual,zhang2017ppr}. 

This task is closely related to object detection, which is a long-standing topic of active research.
%
The goal of object detection is to recognize and localize all instances of a given set of
semantic categories in an image.
Contemporary object detection models employ deep convolutional neural 
networks that are trained on large annotated datasets.
Importantly, many of these models have well-optimized reference implementations,
which work well for a wide variety of object detection datasets,
can be freely accessed and seamlessly deployed into production systems.

Naturally, currently available models for detecting visual relationships~\cite{gkioxari2017detecting,lu2016visual,dai2017detecting} heavily rely on object detection pipelines. 
However, in order to enable the modeling of pairwise relationships, they augment the object detection pipelines with multiple additional components and thereby introduce
additional hyperparameters.
In contrast, in this paper we present a new model that almost exclusively relies on readily available detection pipelines.
Crucially, our model does not require tuning any additional hyperparameters and can be implemented
by adding just a dozen lines of code to existing codebases of widely adopted object object detection pipelines~\cite{lin17iccv,huang2017speed}.
This way researchers and practitioners can fully leverage all advantages of modern object detection systems for solving the task of detecting visual relationships.

We formulate the task of detecting visual relationships as a joint probabilistic model.
Our key idea is to decompose the probabilistic model into two simpler sub-models using the chain rule.
As a result of this decomposition the task of detecting visual relationships breaks down into two consecutive object detection tasks.
A first detection model localizes all objects in an input image.
Then, for each detected object, the second model detects all other objects interacting with it.
Our main contribution is the \emph{Box Attention} mechanism that augments the second model with the ability to be conditioned on objects localized
by the first one.
Specifically, we propose to represent objects detected by the first model as spatial binary masks
encoding their locations. These masks are then given as additional inputs to the second model.
In Section~\ref{sec:method} we explain our approach in detail, and show how to unify the two models into a single unified detection model.

We evaluate our approach on the \VCOCO~\cite{gupta2015visual}, 
\VRD~\cite{lu2016visual} and \OID~\cite{oid18arxiv} datasets in Section~\ref{sec:experiments}.
We show that our model produces highly competitive results on all datasets.
In particular, we match the state-of-the-art results on the \VRD dataset.
On the \VCOCO dataset we achieve the second best result, while only being worse
than a much more complex technique specialized for detecting human-object interactions.
Finally, on the \emph{kaggle} competition\footnote{\url{https://www.kaggle.com/c/google-ai-open-images-visual-relationship-track}} on detecting visual relationships that is based on the \OID dataset, we outperform all techniques that conduct experiments in a  comparable setting and, moreover, achieve the second best absolute place.

%% file: rw.tex
\section{Related Work}

\noindent\textbf{Object detection} pipelines serve as an integral building block for detecting visual relationships.
They take an image as input and output a scored list of object detections
(bounding boxes) labeled by their corresponding semantic classes.
Current state-of-the-art models for object detection achieve impressive performance by leveraging
deep convolutional neural networks.
Our model relies on the RetinaNet detection pipeline~\cite{lin17iccv}.

\noindent\textbf{Visual relationships} have been previously studied in the computer vision community.
Earlier works leverage visual relationships in order to improve performance of object
detection~\cite{sadeghi2011recognition}, action recognition and pose estimation~\cite{desai2012detecting}, or
semantic image segmentation~\cite{gupta2008beyond}.
However, \cite{lu2016visual} was the first work to formulate detection of visual relationships
as a separate task.
It proposes to learn a composite likelihood function that utilizes a language prior based on the word embeddings~\cite{mikolov2013efficient} for scoring visual relationships.
Their inference procedure requires a costly search for high-scoring relationships among all pairs
of objects and all possible relationships between them.
Subsequently, several works improved and generalized that initially proposed model.
In~\cite{yu2017visual} the authors use external sources of linguistic knowledge
and the distillation technique~\cite{hinton2015distilling}
to improve modeling performance, while \cite{liang2017deep} formulates the task
of detecting visual relationships as a reinforcement learning problem.
Dai et al.~\cite{dai2017detecting} propose a multistage relationship detection process,
where they first run an object detector, and then apply a light-weight network for selecting promising pairs of interacting object detections. After that they use an additional neural network to produce relationship predictions.
A similar approach was also leveraged by~\cite{zhang2017visual}.
The work~\cite{li2017vip} introduces a triplet proposal mechanism and then trains a multi-stage scoring function to select the best proposal.
A few works~\cite{peyre2017weakly,zhang2017ppr} investigate a weakly-supervised variant of the relationship detection task.
Further, several works focus on human-object relations. Earlier works exploit probabilistic graphical models~\cite{gupta2009observing,yao2010modeling}
and also investigate weakly-supervised settings~\cite{prest2012weakly}.
Recently, high-performing models~\cite{gupta2015visual,gkioxari2017detecting,gao2018ican} based on deep convolutional neural networks have emerged.

\noindent\textbf{Attention mechanisms.} 
The idea of \textit{modeling structured outputs using attention to earlier predictions} was previously successfully used for a diverse range of computer vision and natural language processing tasks.
Some prominent examples include models for machine translation~\cite{sutskever2014sequence},
image captioning~\cite{vinyals2015show}, speech recognition~\cite{chan2016listen} and
human pose estimation~\cite{gkioxari2016chained}.
Moreover, this idea was also proven to be successful for deriving state-of-the-art generative models
for images~\cite{van2016pixel}, videos~\cite{kalchbrenner2017video} or audio~\cite{oord2016wave}.

Finally, very recent paper~\cite{krishna2018referring} proposes to use an attention mechanism for solving related task of \emph{referring relationships}.
Besides solving a different task, it also uses attention mechanism in a more complex manner.
In contrast, we propose well-justified training procedure based on maximum likelihood principle and provide exact inference algorithm.

%% file: method.tex
\section{Box Attention for Detecting Relationships}\label{sec:method}

In this section we describe our approach for detecting visual relationships, which we call \method (\emph{\textbf{B}ox \textbf{A}ttention \textbf{R}elational CNN}).
We first give a high-level overview in Section~\ref{subsec:overview} and then present a detailed description of its core elements in Section~\ref{subsec:model_details}. Finally, we discuss properties of our model in Section~\ref{subsec:discussion}.

\begin{figure*}
    \centering
    \includegraphics[width=1.0\textwidth]{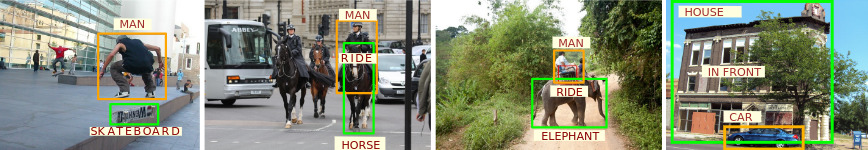}
    \caption{Examples of interacting objects.
             Subjects and objects are annotated by orange and green boxes respectively.
             Green boxes are also marked by labels that describe relationships to the corresponding subjects.}
    \label{fig:examples}
\end{figure*}

\subsection{Overview}
\label{subsec:overview}

Our overall approach is illustrated in Fig~\ref{fig:method}.
Formally, the task of detecting visual relationships for a given image can be formulated as
detecting all triplets in a form of \emph{$\langle$subject ($S$), predicate ($P$), object ($O$)$\rangle$}.
The subject $S$ and object $O$ are represented by bounding boxes $b^s$ and $b^o$, and their corresponding category labels by $l^s$ and $l^o$. 
The predicate $P$ is represented by a label $l^p$.
A single subject may be in a relationship with multiple objects and a single object may by in a relationship with multiple subjects. 
Some examples of such interactions are illustrated in Figure~\ref{fig:examples}.

We derive our approach for modeling visual relationships using a probabilistic interpretation of this task.
The high-level idea is to model the probability $\p(S, P, O | I)$ that
a triplet $\langle S, P, O \rangle$ is a correct visual relationship in the input image $I$.
It is challenging to model this joint probability distribution, as it involves multiple
structured variables interacting in a complex manner.
Thus, we propose to employ the chain rule in order to decompose the joint probability into simpler conditional probabilities:
\begin{equation}
    \p (S, P, O | I) = \p(S | I) \cdot \p( P, O | S, I). \label{eq:model}
\end{equation}

The first factor $\p(S | I)$ models the probability that a subject $(b^s, l^s)$ is present in the image $I$. 
Thus, this factor can be modeled as a standard detection task of predicting bounding boxes and category labels for all instances in the image.
The second factor,  $\p( P, O | S, I)$, models the probability that an object $(b^o, l^o)$ is present in the image and is related to the subject $S$ through a predicate $l^p$.
%
Estimating $\p( P, O | S, I)$ can be also seen as a detection problem. In this case the model should output bounding boxes, object labels and the corresponding predicate labels $(b^o, l^o, l^p)$ for all objects that interact with $S$.
We implement conditioning on $S$ by treating it as an additional input that
we call \emph{Box Attention}.
In Section~\ref{subsec:model_details} we present this \emph{Box Attention} mechanism in detail.

Due to functional similarity of the two factors in Eq.~\eqref{eq:model} we further propose to train a single unified model for both $\p(S | I)$ and $\p( P, O | S, I)$. 
Note that our approach can be implemented within any object detection model.
From now on we will refer to it as the {\em base detection model}.
Below we provide a more detailed description of our model as well as inference and training procedures.

\subsection{Model details}
\label{subsec:model_details}

\paragraph{Box attention representation.}
Consider an input image $I$.
The box attention map for this image is represented
as a binary image $m$ of the same size as $I$, with $3$ channels.
The first channel represents a subject bounding box (Figure~\ref{fig:method}).
Specifically, all pixels inside the subject bounding box are set to $1$ and all other pixels are set to $0$.
An attention map can be empty: in this case the first channel is all zeros.
The second and third channels are used in the following way: 
if the first channel is not empty, then the second channel is all zeros and the third channel is all ones.
Conversely, if the first channel is empty, then the second channel
is all ones and the third channel is all zeros.
These two extra channels are useful because state-of-the-art detection models use deep convolutional neural networks as feature extractors~\cite{he16cvpr,szegedy2016rethinking}.
Neurons of these networks have limited receptive fields that might not cover the whole attention map.
As a consequence, these neurons have no information whether the attention map is empty or not based only on the first channel.

\paragraph{Incorporating box attention maps in the base detection model.}
In order to incorporate the additional box attention input we use
a simple and yet very effective strategy.
It is inspired by approaches for
conditioning probabilistic models on external inputs
from the literature on generative image modeling~\cite{oord2016conditional,salimans2017pixelcnn++}.

\begin{figure}
\centering
\includegraphics[width=0.25\textwidth]{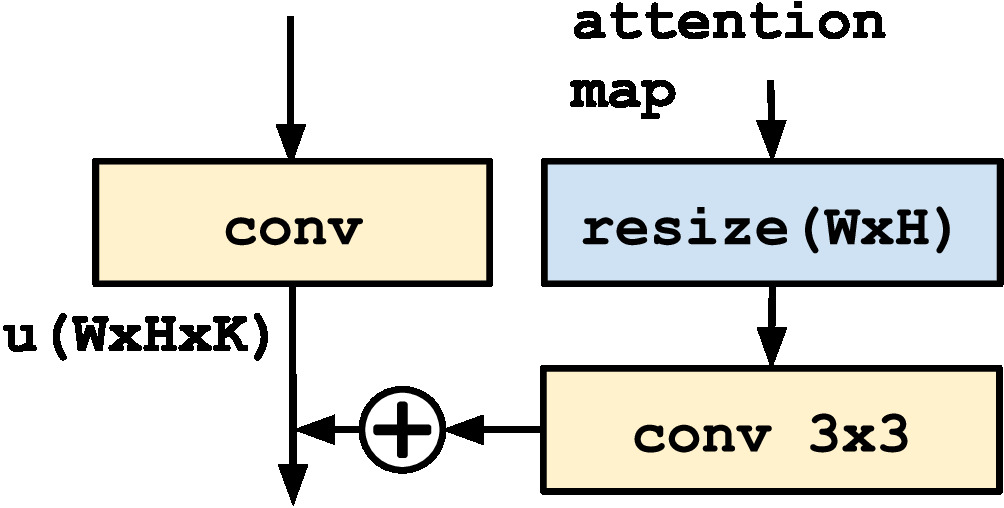}
\caption{Our proposed attention mechanism. This procedure is applied to convolutional layers of the base detection model.}\label{fig:attention}
\end{figure}

The proposed mechanism is illustrated in Figure~\ref{fig:attention}.
Consider the output $u$ of a certain convolutional layer of the base detection model.
Let's assume $u$ has spatial resolution $H \times W$ and $K$ channels.
We condition the output $u$ on the attention map $m$ by performing the following steps:
\begin{enumerate}
\item Obtain $\hat{m}$ by resizing $m$ to the spatial size of $H \times W$
      using nearest neighbor interpolation.
\item Obtain $\tilde{m}$ by passing $\hat{m}$ through a learnable convolutional layer with $K$ output channels and a kernel of size $3 \times 3$.
\item Update $u$ as  $u + \tilde{m}$.
\end{enumerate}
In principle, we can apply this procedure to every convolutional layer of the base detection model.
In practice, we use ResNet-type architectures with bottleneck units~\cite{he16cvpr}, and apply the above conditioning procedure to the second convolution of every bottleneck unit.

The proposed conditioning procedure has several appealing properties.
First, it allows to seamlessly initialize the \method model using the pre-trained base detection model.
Second, if we initialize the convolutional kernels in step 2 with all zeros, in the beginning of training our conditioning procedure does not have any effect on the outputs of the base detection model.
This helps preventing disruption of the pre-trained base detection model
and ensures numerical stability in the beginning of the training process.
Finally, the proposed conditioning is easy to implement as it
only requires modifying a single function that creates convolutional layers.

\noindent\textbf{Training.}
As described in Section~\ref{subsec:overview}, we want to learn a single model able to a) output subject predictions when the attention map is empty, and
b) output object and predicate predictions when conditioned on a subject prediction through the attention map.

To achieve a) we simply add an empty attention map to each image in the training set and preserve all subject bounding boxes as ground-truth. This forms one type of training sample.

To achieve b) for each subject annotated in a training image we generate a separate training sample consisting of the same image, box attention map corresponding to the subject, and all object and predicate annotations corresponding to this subject. This is the second type of training sample.
%
Thus, given $k$ annotated subjects in a training image, we create $k+1$ training samples.
We then use this training set to train a \method model. 
Note that storing multiple copies of the image is not necessary, as the training samples can be generated on the fly during the course of training.

According to the \method formulation in Section~\ref{subsec:overview}, the model predicts two labels (i.e. $l^o$ and $l^p$) per detection instead of one.
Therefore during training we use a sigmoid multiclass loss instead of a cross-entropy loss normally used for standard object detection.
During training we closely follow recommendations from the public RetinaNet implementation\footnote{https://github.com/tensorflow/tpu/tree/master/models/official/retinanet} (see Section~\ref{sec:experiments} for more details).

\paragraph{Predicting visual relationships.}

\begin{figure}
\centering
\includegraphics[width=0.4\textwidth]{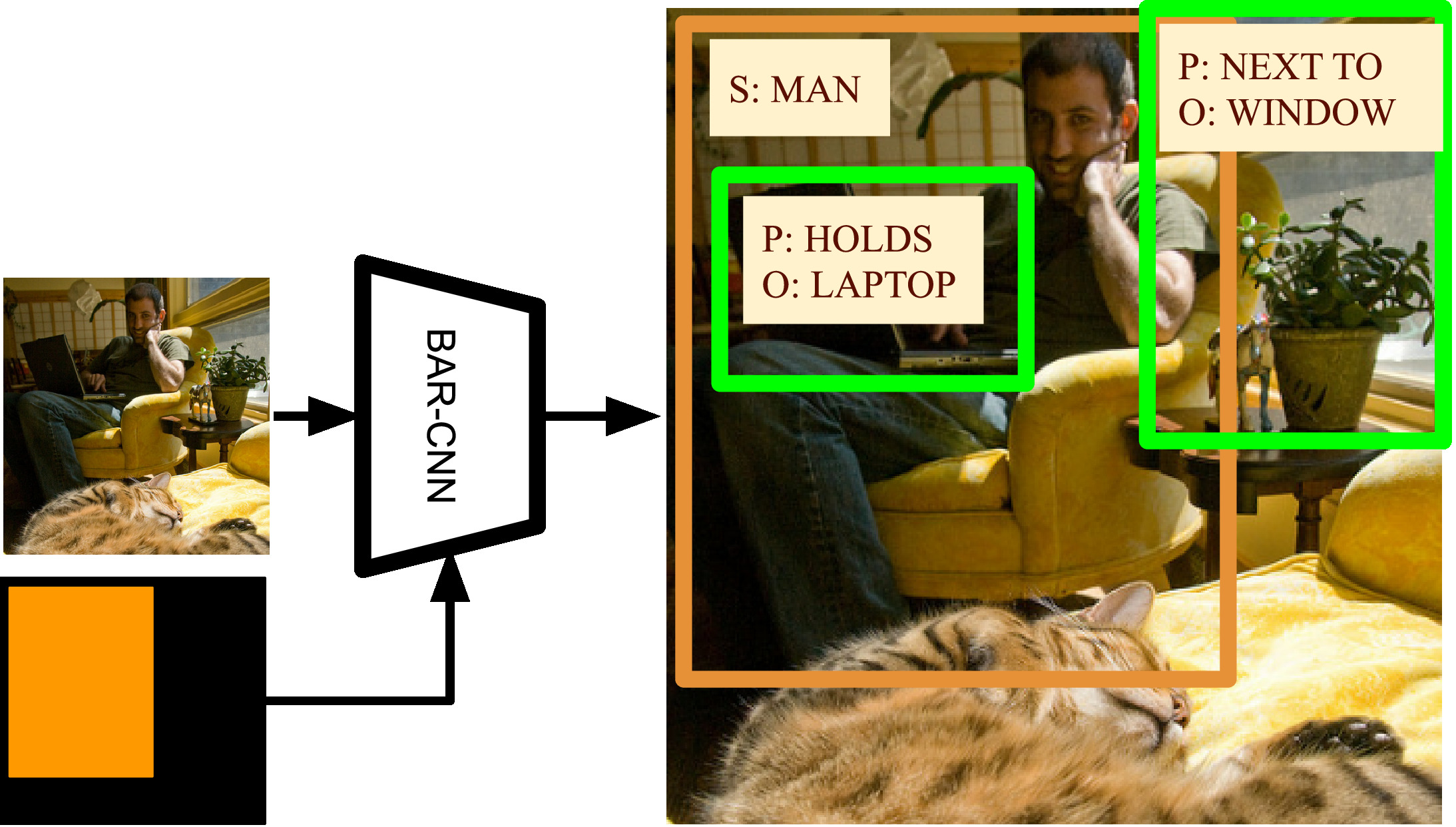}
\caption{Example of the inference procedure: first, attention mask with the subject (in orange) and image is passed through the model. The model makes two predictions, related to this subject (in green). }\label{fig:inference}
\end{figure}

Visual relationships for a test image $I$ are predicted using a natural two-stage procedure.
First, we run the \method
model by feeding it the image $I$ and an empty attention map.
The model outputs a scored list of subject detections. Each detection has a box $b^s$, a class label $l^s$ and a score $s^s$.

Second, we form a scored list of all detected relationships.
Specifically, for every detection we apply the following procedure:
\begin{enumerate}

\item Construct an attention map for $b^s$ and feed it to the
       \method model together with the image $I$. As a result \method predicts a set of object boxes $(b^o, l^o, l^p)_i$ that interact with $b^s$ through the relationship $l^p$.
\item For every detection $(b^o, l^o, l^p)$ we first compute  the score $\p(P,O|S,I) = s^{p,o}$ by multiplying the scores of labels $l^o$ and $l^p$ obtained through multiclass prediction.
The final score $s$ of the full visual relationship detection $\langle(b^s, l^s), l^p, (b^o, l^o)\rangle$ is computed as $s = s^s s^{p,o}$.

\end{enumerate}

The process is illustraited in the Figure~\ref{fig:inference}.

\subsection{Discussion.}
\label{subsec:discussion}

In this section we proposed a conceptually clean and powerful approach for detecting visual relationships.
Indeed, our model is formulated as end-to-end optimization
of a single detection model with a theoretically sound objective.
Further, the predictions scoring has a justified probabilistic interpretation~\eqref{eq:model}.

Importantly, unlike some previously proposed models,
\method does not make any 
restrictive modeling assumptions (except for those already present in the base detection model).
For instance,~\cite{gkioxari2017detecting} assumes
that the distribution of object box location conditioned on a given subject box and action can be modeled by a Gaussian distribution.
Another interesting example is~\cite{li2017vip}.
It proposes to generate relationship proposals and learn a multi-stage model to select the most promising candidates.
However, since the number of initial proposals is huge (due to combinatorial explosion),~\cite{li2017vip} employs a heuristical filtering mechanism that is not guaranteed to work in the general case.

%% file: experiments.tex
\section{Experiments}\label{sec:experiments}

We now present experimental evaluation of the proposed \method model. 
We evaluate on the three publicly available datasets:
\VRD~\cite{lu2016visual}, \VCOCO~\cite{gupta2015visual} and \OID~\cite{oid18arxiv}\footnote{For evaluation we use the Open Images Challenge 2018 evaluation server.}, reporting strong quantitative and qualitative results.

\subsection{Implementation details}\label{subsec:impl_details}
As explained in Section~\ref{sec:method},
we build \method by combining the base detection model with the box attention input.
In our experiments we use the RetinaNet~\cite{lin17iccv} model with ResNet50~\cite{he16cvpr} backbone as the base detection model.

During finetuning we do not freeze any of the network's parameters.
As an optimization algorithm we use stochastic gradient descent
with momentum set to 0.9 and
the mini-batch size is set to 256 images.

Before finetuning on the \VCOCO~and \VRD~datasets, we initialize a \method model from the \emph{RetinaNet} detection model pretrained on the \emph{MSCOCO} \emph{train2014} split.
For finetuning on the \OID~dataset, we initialize our model from the \emph{RetinaNet} detection model pretrained on bounding boxes of the training split of the \OID~dataset itself.
We conduct finetuning for 60, 30 and 15 epochs for the \VCOCO, \VRD~and \OID~datasets, respectively. 
The initial learning rate is always set to $8 \cdot 10^{-3}$ and is decayed twice by a factor of 10 after 50\% and 75\% of all optimization steps.
All other hyperparameters of the \emph{RetinaNet} model are set to their default values from the publicly available implementation\footnote{\url{https://github.com/tensorflow/tpu/tree/master/models/official/retinanet}}.

\begin{figure*}[t]
\setlength{\fboxsep}{0pt}
\resizebox{\linewidth}{!}{%
 \includegraphics[height=2cm]{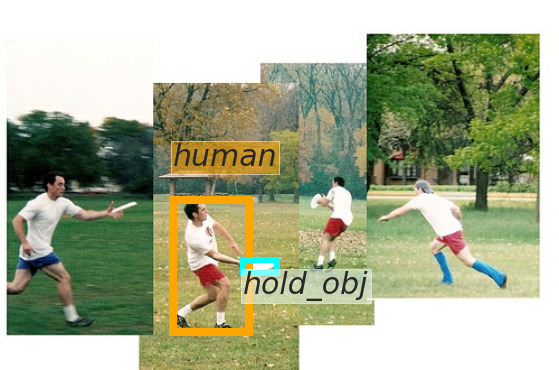}
 \includegraphics[height=2cm]{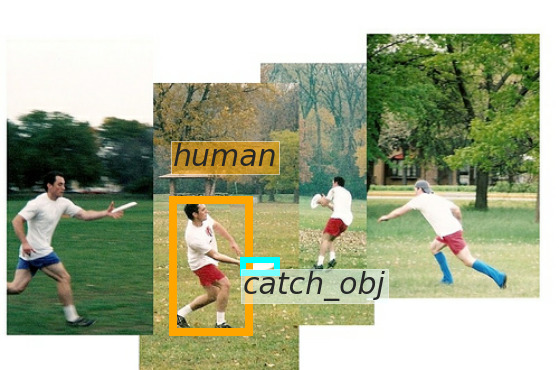} 
 \includegraphics[height=2cm]{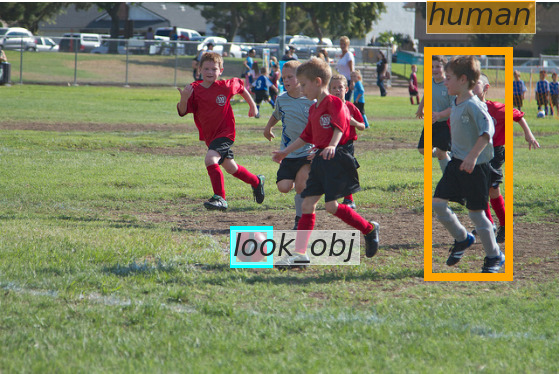}
 \includegraphics[height=2cm]{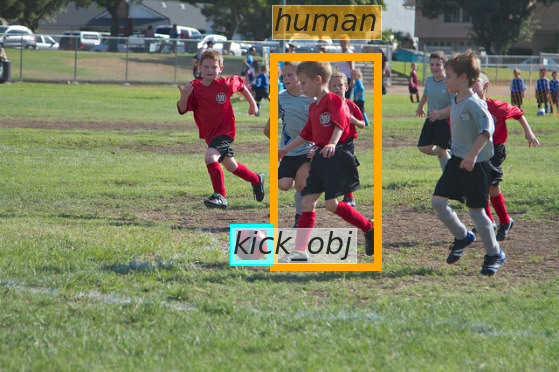}
 \includegraphics[height=2cm]{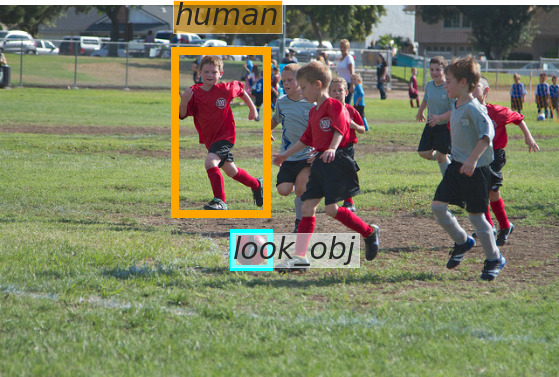} 
}
\\[1mm]
\resizebox{\linewidth}{!}{%
 \includegraphics[height=2cm]{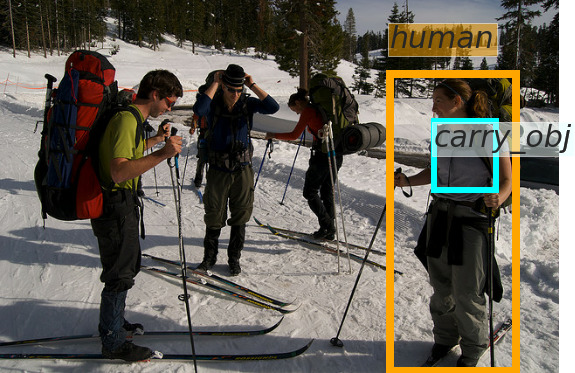}
 \includegraphics[height=2cm]{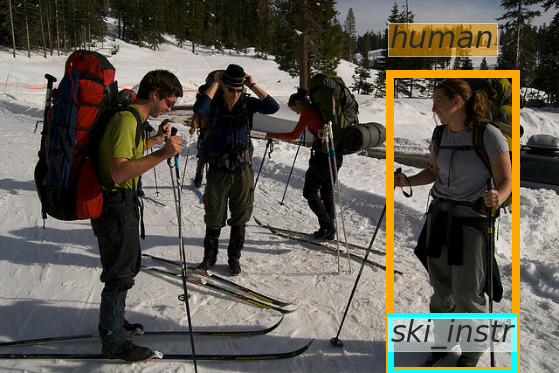}
 \includegraphics[height=2cm]{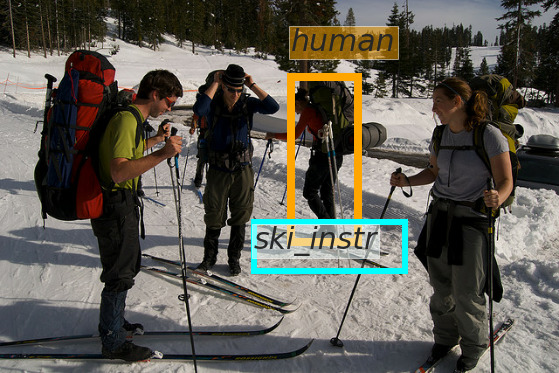}
 \includegraphics[height=2cm]{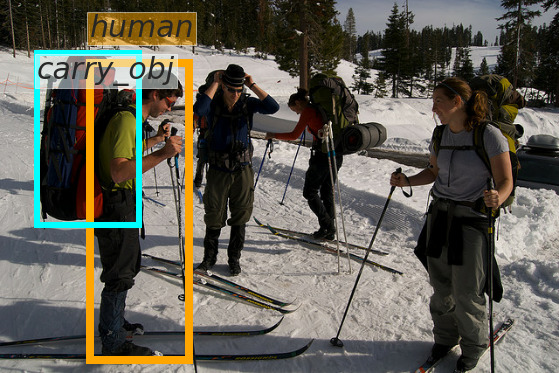}
 \includegraphics[height=2cm]{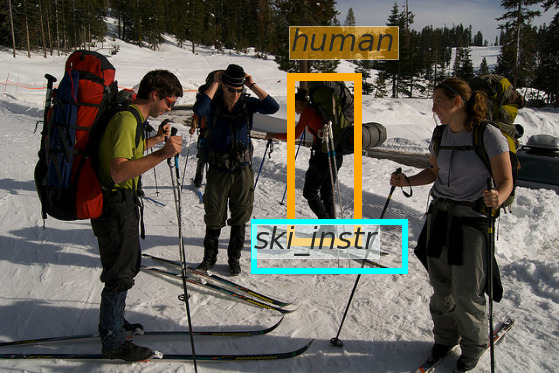} 
}
\\[3mm]
\resizebox{\linewidth}{!}{%
 \includegraphics[height=2cm]{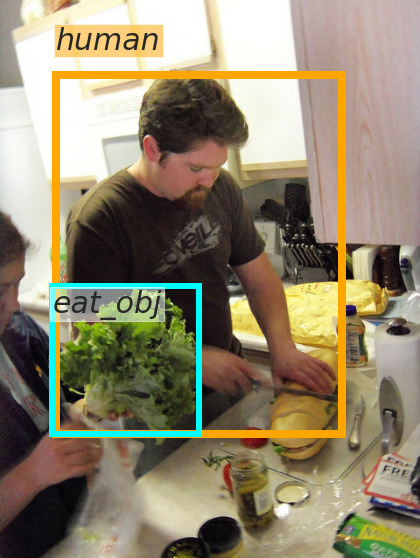}
 \includegraphics[height=2cm]{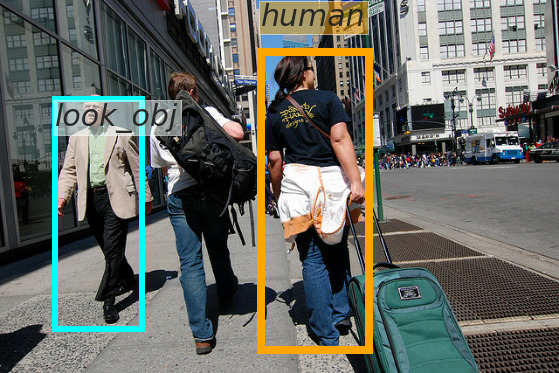}
 \includegraphics[height=2cm]{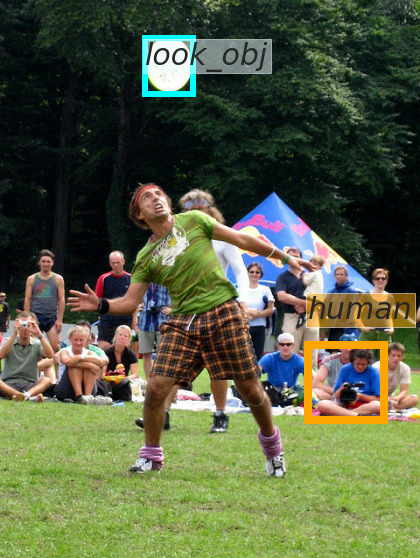} 
 \includegraphics[height=2cm]{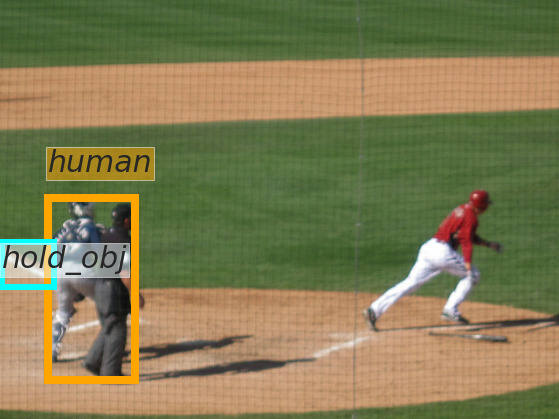} 
 \includegraphics[height=2cm]{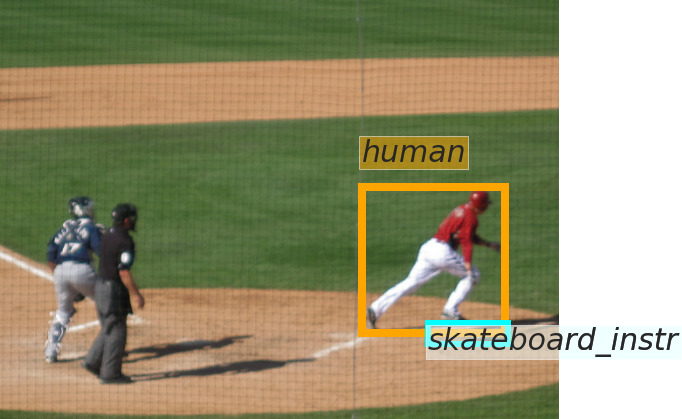} 
}
    \caption{\small Example outputs of the top scoring detections by the proposed \method model on \VCOCO.
    The first $2$ rows demonstrate correct predictions outputted by the model. The $3$d row shows failures: from left to right: image $1,2,3$ --- wrong target, images $4,5$ --- hallucinated object.}
    \label{fig:qualitative_vcoco}
\end{figure*}

\newcommand{\midd}[1]{\multirow{2}{*}{{#1}}}
\newcommand{\objinstr}{\begin{tabular}[l]{@{}l@{}}{\footnotesize ({\it object})} \\ {\footnotesize ({\it instr.})}\end{tabular}}
\begin{table}[t]\center
\begin{tabular}{|c||c|c|c|}
\hline
 & Model C~\cite{gupta2015visual}
 & InteractNet
 & \method
 \\
 & impl. \cite{gkioxari2017detecting}
 & \cite{gkioxari2017detecting}
 & (proposed) \\
\hline
 Relation & \textbf{AP$_\text{role}$}
          & \textbf{AP$_\text{role}$}
          & \textbf{AP$_\text{role}$}
 \\
\hline
 carry & 8.1  & 33.1 & 31.0 \\
 catch & 37.0 & 42.5 & 34.6 \\
 drink & 18.1 & 33.8 & 33.1 \\
 hold  & 4.0  & 26.4 & 40.3 \\
 jump  & 40.6 & 45.1 & 61.1 \\
 kick  & 67.9 & 69.4 & 67.2 \\
 lay   & 17.8 & 21.0 & 34.3 \\
 look  & 2.8  & 20.2 & 32.7 \\
 read  & 23.3 & 23.9 & 24.1 \\
 ride  & 55.3 & 55.2 & 55.9 \\
 sit   & 15.6 & 19.9 & 34.7 \\
 skateboard & 74.0 & 75.5 & 76.6 \\
 ski   & 29.7 & 36.5 & 33.3 \\
 snowboard  & 52.8 & 63.9 & 52.7 \\
 surf & 50.2 & 65.7 & 57.7 \\
 talk-phone & 23.0 & 31.8 & 40.0 \\
 throw & 36.0 & 40.4 & 29.5 \\
 work-comp & 46.1 & 57.3 & 57.6 \\
\hline
\midd{cut \, \objinstr} & 16.5 & 23.0 & 33.3 \\
                        & 15.1 & 36.4 & 23.2 \\
\midd{eat \, \objinstr} & 26.5 & 32.4 & 44.3 \\
                        & 2.7  & 2.0  & 53.5 \\
\midd{hit \, \objinstr} & 56.7 & 62.3 & 29.5 \\
                        & 42.4 & 43.3 & 65.4 \\
\hline
mean AP & 31.8 & 40.0 & \textbf{43.6} \\
\hline
\end{tabular}
\caption{\small Quantitative comparison of the proposed model with competing models on \VCOCO.}\label{table:res_vcoco_detailed}
\end{table}

\subsection{Results on the VCOCO dataset~\cite{gupta2015visual}}

\noindent\textbf{Data.}
The V-COCO dataset contains natural images annotated by human-object
relationships.
There are 29 relationships (also called actions), \eg
\emph{carry, drink, ride, cut, eat (object), eat (instrument), \etc}
Following previous work~\cite{gupta2015visual,gkioxari2017detecting},
we drop 5 actions from the evaluation procedure:
\emph{run, smile, stand, walk} because they do not have target objects,
and \emph{point} because it appears very rarely (only 31 instances in the test set).
Overall, the dataset has $5,400$ images in the
joint \emph{train} and \emph{val} splits, and
$4,946$ images in the \emph{test} split.
On average each image has $4.5$ annotated human-object relationships.

The ground-truth relationships may contain objects marked as \emph{not visible}. 
As an example, imagine \emph{a human sitting on a chair}, where the chair is \emph{not visible} because it is hidden by a table.
Thus, in our training procedure we also consider a special case to handle this situation.
In this case we augment a human bounding box with an extra label indicating that the "sitting" predicate has an object being not visible.

Note that the \emph{test} split contains only images from the COCO \emph{val} split,
which were not used for pre-training the base detection model.

\noindent\textbf{Metric.}
We evaluate performance on the V-COCO dataset using its official metric: ``AP role''~\cite{gupta2015visual}.
This metric computes the mean average precision (mAP) of detecting relationships triplets
\emph{$\langle$human box, action, target box$\rangle$} by
following the PASCAL VOC 2012~\cite{everingham2015pascal}
evaluation protocol.
A triplet prediction is considered as correct, if all three of its components are correct.
A predicted box is correct if it has intersection-over-union with
a ground-truth box of at least $50\%$.
We use the official publicly available code for computing
this metric\footnote{\url{https://github.com/s-gupta/v-coco}}.

\noindent\textbf{Qualitative results.}
Figure~\ref{fig:qualitative_vcoco} show typical outputs of our model.
By analyzing these qualitative results, we make a few observations.
Most importantly, our model successfully learns to use the box attention map.
Even for complex images with many objects it learns to correctly assign humans to their
corresponding objects.
For instance, row $2$ shows four humans, each correctly assigned to the objects they interact with.
We also note that our model can predict multiple objects interacting with a single human through different relationships (row $2$, two left-most images).
Interestingly, \method can also successfully predict that the same object
can correspond to different humans through different actions (row $1$, the right-most three images).
Moreover, \method can model long range interaction between objects,
\eg in row $1$ the two football players looking at the ball are far from a ball and yet are predicted to be related to it through the action \emph{look}.

In the V-COCO dataset most errors are caused by complex image semantics,
which are hard to capture by a neural network trained on very limited amount of data (row $3$).
Another interesting type of error is object hallucination.
For instance, \method learned that if a human stands in a specific pose he is skateboarding. However, this is not always the case, as can be seen in the last image of row $3$.
Another common failure is to assign the wrong target object --- for example in the left-most image in row $3$ the \method incorrectly predicts that a human is eating the salad wheres in the image somebody else is holding the salad.

\noindent\textbf{Quantitative results.} 
Results are presented in Table~\ref{table:res_vcoco_detailed}.
The first two columns show quantitative comparison to the model from~\cite{gupta2015visual}
and the approach from~\cite{gkioxari2017detecting}.
Our method \method (third column) outperforms both of them.

The recently proposed ICAN model~\cite{gao2018ican} does not report per-class AP, so we do not include it in Table~\ref{table:res_vcoco_detailed}.
It achieves a $45.3$ mean AP, which is slightly better than our model ($43.6$).
However, we stress that
(1) our model handles the generic visual relationship detection task, whereas ICAN focuses on human-object interaction;
(2) the ICAN model is much more complex than ours, as it introduces numerous additional components on top of the object detection pipeline.
Importantly, our model does not require tuning of any hyperparameters, as they are all inherited from the base detection model and we keep them to their default values.

\subsection{Results on the Visual Relationship dataset~\cite{lu2016visual}}

\begin{table}[t]\center
\begin{tabular}{|c|c|}
\hline
 & Recall@50/100 \\
\hline 
V+L+K \cite{lu2016visual} & $13.86$ / $14.70$ \\
VTransE \cite{zhang2017visual} & $14.07$ / $15.20$ \\
DR-Net \cite{dai2017detecting} & $17.73$ / $20.88$ \\
ViP-CNN \cite{li2017vip} & $17.32$ / $20.01$ \\
VRL \cite{liang2017deep} & \bf{18.19} / $20.79$ \\
\hline
\method (proposed) & 17.75 / \bf{21.60} \\
\hline
\end{tabular}
\caption{\small Quantitative comparison of the proposed model with competing models on \VRD.}\label{table:res_vrd}
\end{table}

\begin{figure*}[t]
\setlength{\fboxsep}{0pt}
\resizebox{\linewidth}{!}{%
    \includegraphics[height=2cm]{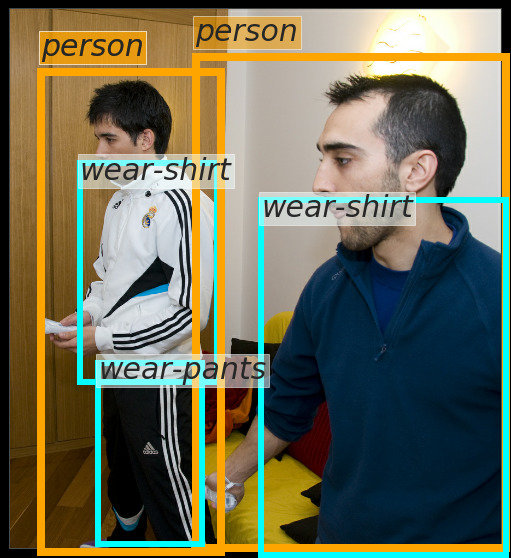}
    \includegraphics[height=2cm]{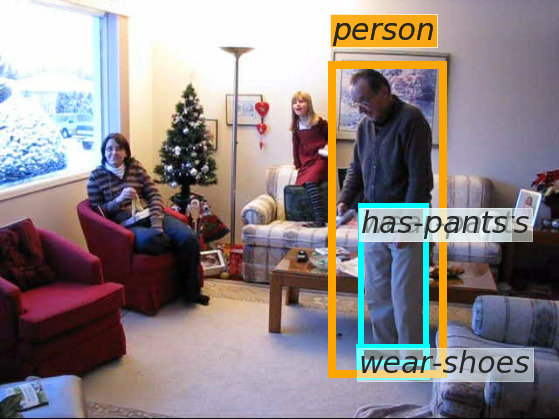}
    \includegraphics[height=2cm]{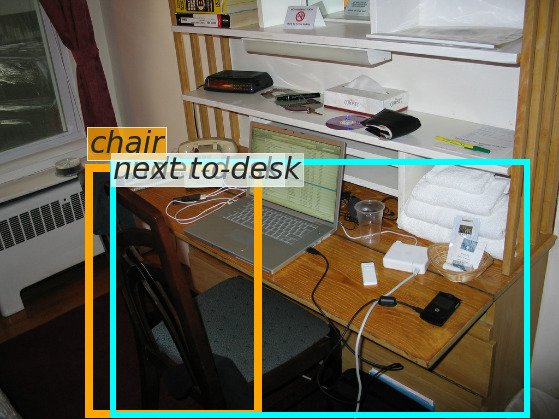}
    \includegraphics[height=2cm]{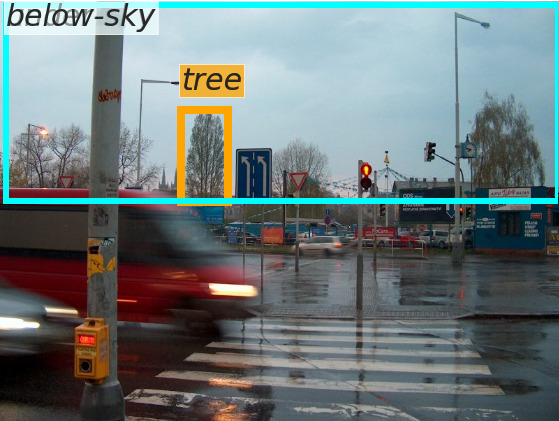}  
    \includegraphics[height=2cm]{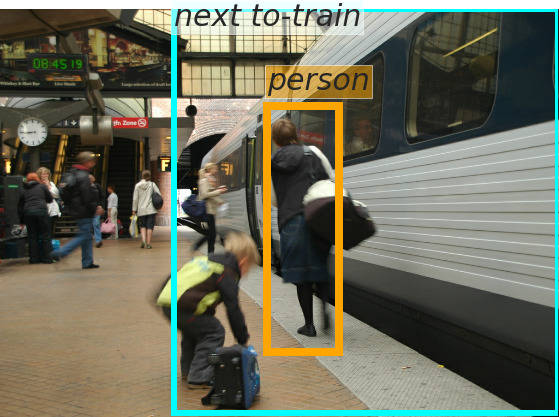}     
    } 
    \\[1mm]
\resizebox{\linewidth}{!}{%
    \includegraphics[height=2cm]{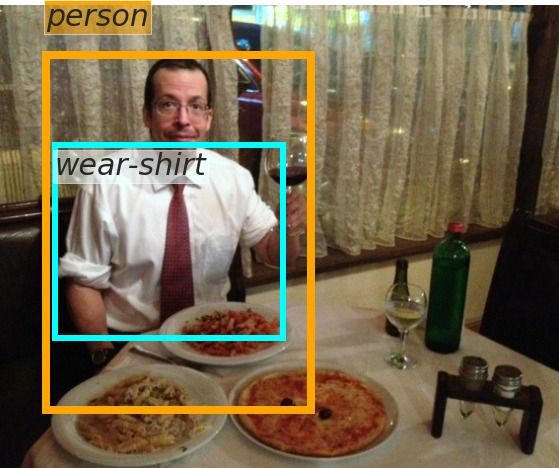}
    \includegraphics[height=2cm]{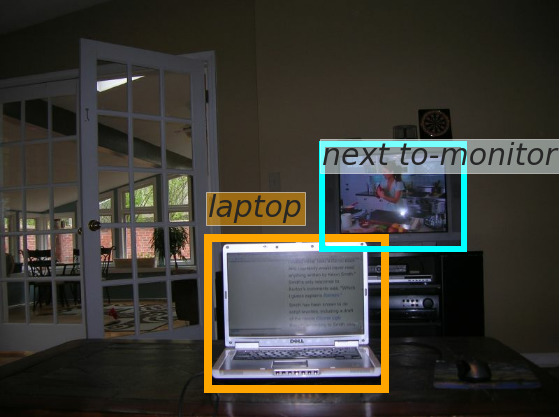}
    \includegraphics[height=2cm]{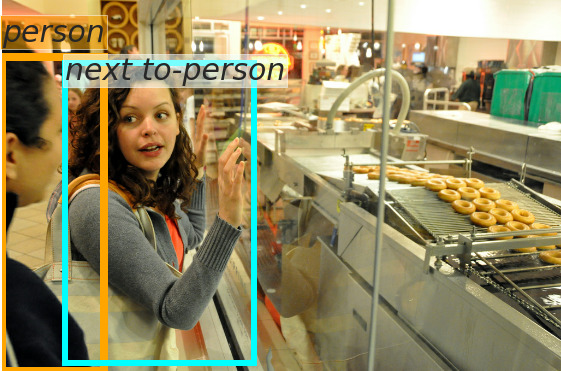}
    \includegraphics[height=2cm]{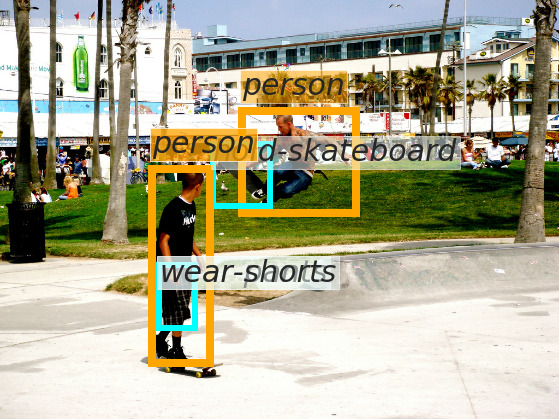}
    \includegraphics[height=2cm]{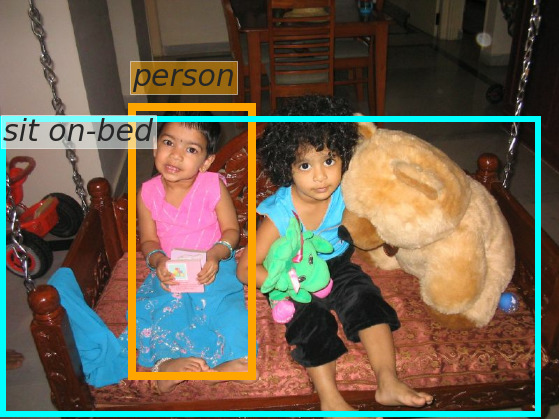}     
    }
    \\[3mm]
\resizebox{\linewidth}{!}{%
    \includegraphics[height=2cm]{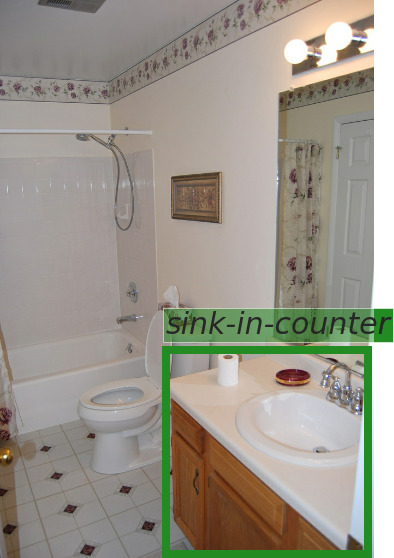}
    \includegraphics[height=2cm]{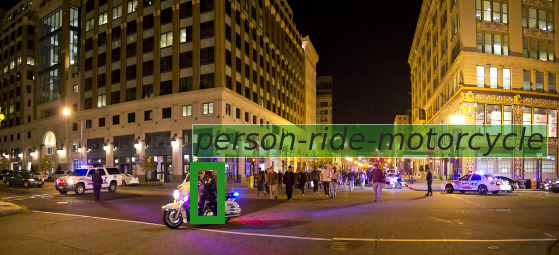}    
    \includegraphics[height=2cm]{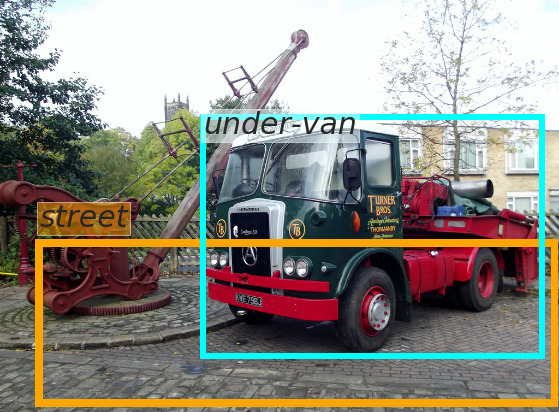}
    \includegraphics[height=2cm]{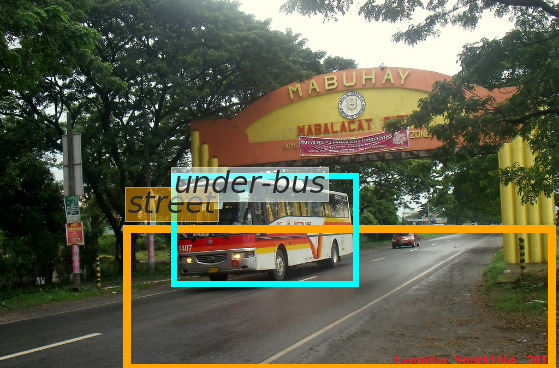} 
    \includegraphics[height=2cm]{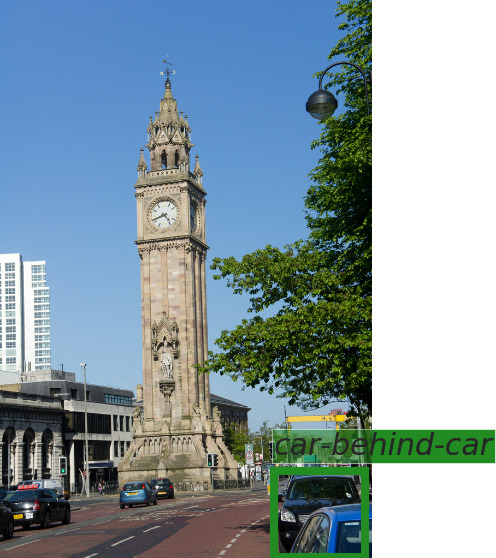}     
    }    
    \caption{\small Example outputs of the top scoring detections by the proposed \method model on \VRD.
    The first row demonstrate correct top-scoring predictions of the model; the second row shows $5$ top scoring predictions on the same image. The $3$d row shows common failures: from left to right, images $1,2$: the box for the same object is predicted; $3$ --- wrong class label; $4$ wrong box; $5$: the same instance is predicted for the subject and the object.}
    \label{fig:qualitative_vrd}
\end{figure*}

\noindent\textbf{Data.}
This dataset contains images annotated by very diverse relationships between objects, not only human-centric ones.
It has $70$ relationships of various nature,
such as spatial relationships (\eg \emph{behind, next to}),
comparative relationships (\eg \emph{taller than}),
and actions (\eg \emph{kick, pull}).
There are $4000$ training and $1000$ test images.
On average each image is annotated by $7.6$ relationships.

\noindent\textbf{Metric.}
We also use the official metrics for this dataset: \emph{Recall@50} and \emph{Recall@100}~\cite{lu2016visual}
\footnote{\url{https://github.com/Prof-Lu-Cewu/Visual-Relationship-Detection}.}.
These metrics require the model to output $50$ or $100$ relationship detections per image, and measure the 
percentage of ground-truth annotations that are correctly detected by these guesses (\ie measuring recall, 
without considering precision).
The notion of correctness is analogous to that of the ``AP role'' metric from \VCOCO, except that class labels for subject and object are taken into account.

Figure~\ref{fig:qualitative_vrd} shows example results.
These demonstrate that our model can predict generic object-to-object relationships, not only
human-centric ones.

For this dataset we also observe that many errors are caused by the model's inability to fully capture complex semantics.
Moreover, as this dataset has much more diverse objects, many mistakes are caused by poor
object detections.

\noindent\textbf{Quantitative results.}
Table~\ref{table:res_vrd} shows quantitative results for this dataset, which is significantly different from V-COCO.
This dataset has much more diverse relationships of complex nature, not only human-centric ones.
We compare to the model from~\cite{lu2016visual} and to the top-performing methods~\cite{dai2017detecting,li2017vip,liang2017deep,zhang2017visual}.
Our method, \method, matches the multi-component top-performing model \cite{liang2017deep}.

\subsection{Results on Open Images VRD Challenge 2018}

\begin{table}[t]\center
\begin{tabular}{|c|c|c|}
\hline
 & Score(public) & Score(private) \\
\hline 
team MIL & $21.8$ & $19.7$ \\
team mission-pipeline & $7.4$ & $6.8$ \\
team toshif & $25.6$ & $22.8$ \\
\hline
FREQ & $15.2$ & $12.2$ \\
FREQ-OVERLAP & $22.0$ & $16.9$ \\
\hline
\method (proposed) & \bf{26.6} & \bf{25.0} \\
\hline
\end{tabular}
\caption{\small Quantitative comparison of the proposed model with competing models on \OID (all scores are multiplied by $100$ for convenience).  }\label{table:res_oid}
\end{table}

\begin{figure*}[t]
\setlength{\fboxsep}{0pt}
\resizebox{\linewidth}{!}{%
    \includegraphics[height=2cm]{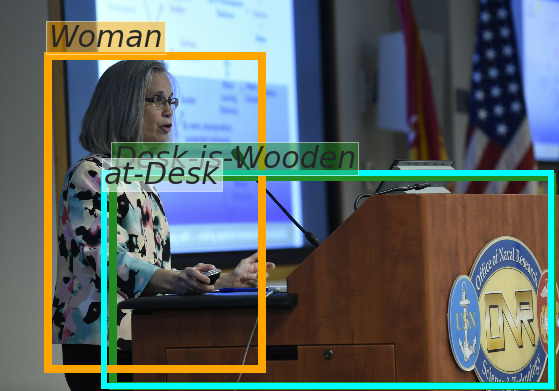}
    \includegraphics[height=2cm]{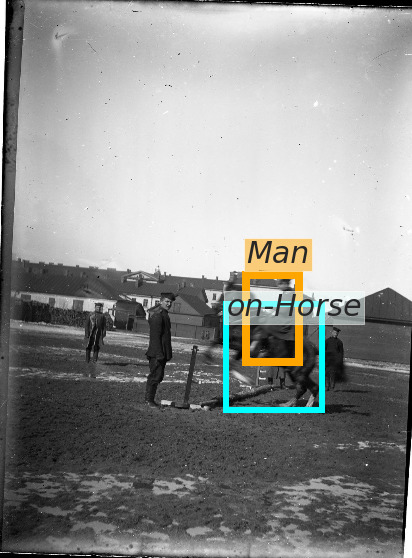}
    \includegraphics[height=2cm]{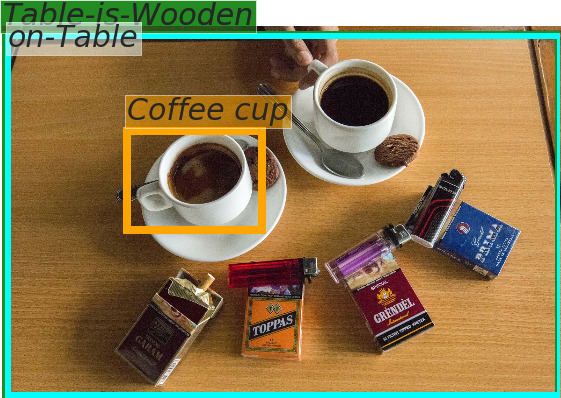}  
    \includegraphics[height=2cm]{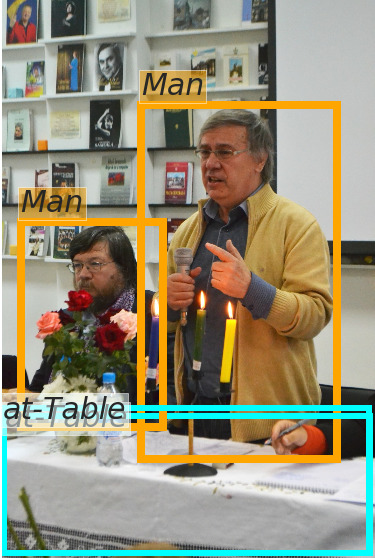} 
    \includegraphics[height=2cm]{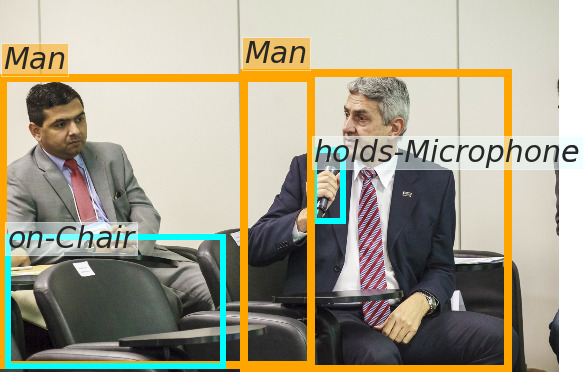}
    } 
    \\[1mm]
\resizebox{\linewidth}{!}{%
    \includegraphics[height=2cm]{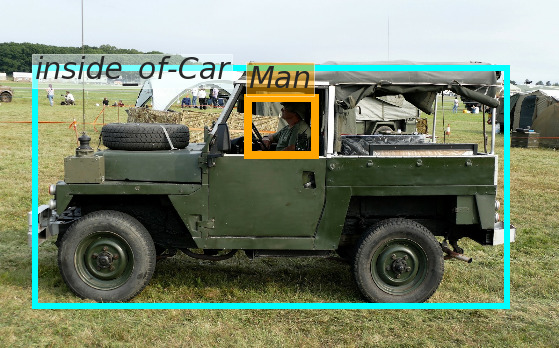}
    \includegraphics[height=2cm]{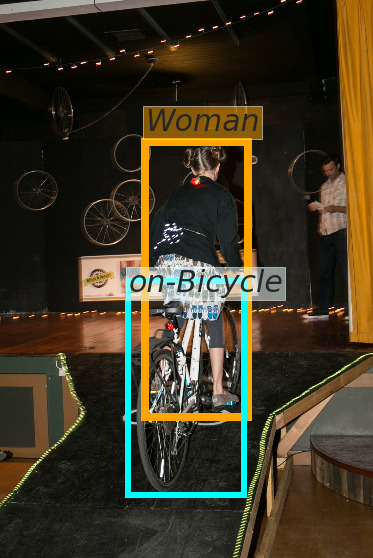}
    \includegraphics[height=2cm]{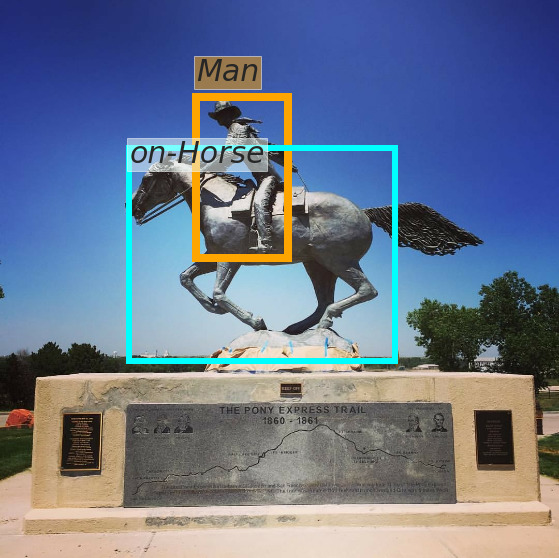}
    \includegraphics[height=2cm]{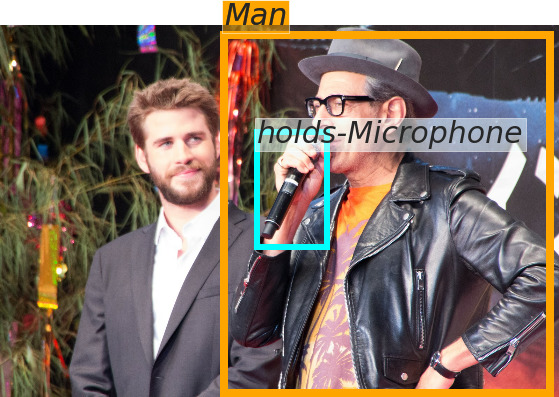}
    \includegraphics[height=2cm]{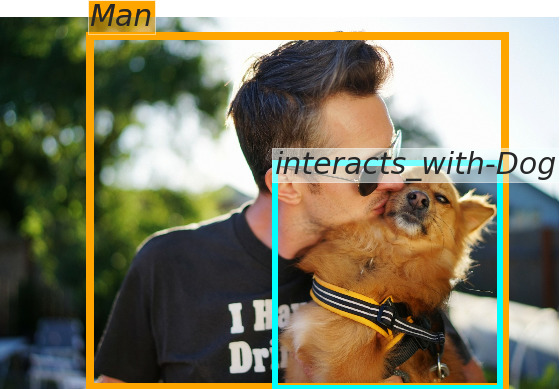}   
    } 
    \caption{\small Example outputs of the proposed \method model on \OID:
    top-scoring predictions on \OID challenge set.}
    \label{fig:qualitative_oid}
\end{figure*}

\noindent\textbf{Data.}
The \OID Dataset (OID) is a very large-scale dataset containing image-level labels, object bounding boxes, and visual relationships annotations.
In total it contains $329$ distinct relationship triplets and $374,768$ annotations on $100,522$ images in the training set.

\noindent\textbf{Metric.}
We evaluate the model on the hidden Open Images Challenge 2018\footnote{ \url{https://storage.googleapis.com/openimages/web/challenge.html}} test set
using the official Kaggle server\footnote{\url{https://www.kaggle.com/c/google-ai-open-images-visual-relationship-track}}.
The metric is the weighted average of the three metrics common for visual relationship detection performance evaluation:
mAP on phrase detection, mAP for relationship detection and Recall@50 for relationship detections\footnote{\url{https://storage.googleapis.com/openimages/web/vrd_detection_metric.html}}. The task of phrase detection is to detect triplets of object with a single enclosing bounding box and three labels $l^s,l^p,l^o$ (as defined in Section~\ref{subsec:overview}).
It was introduced in~\cite{lu2016visual}.
The two other metrics require detecting separately each object and their relationship label.
For the mAP metrics, the mean is computed over relationship predicate, \ie $l^p$.

\noindent\textbf{Qualitative results.}
The top scoring detections of the model are displayed on the Figure~\ref{fig:qualitative_oid}. Open Images is challenging dataset with very diverse and complex scenes. 

A lot of errors we observed are associated with the wrong predictions of the object or subject labels and wrong bounding box. In this sense the mistakes are similar to those on the VRD dataset.

\noindent\textbf{Baselines.}
Since we are not aware of any published  baseline on this dataset, we compute two frequency baselines inspired by~\cite{zellers2018scenegraphs}. As in~\cite{zellers2018scenegraphs} we name them FREQ and FREQ-OVERLAP.

The joint probability distribution in Eq.~\eqref{eq:model} can be further decomposed into:
\begin{equation}
 \p(S, P, O | I) = \p(S| I) \cdot \p(P | O, S, I) \cdot \p(O | S, I)  \label{eq:baselines}   
\end{equation}
In the simplest case $\p(P | O, S, I)$ can be computed from the training set distribution as the prior probability to have a certain relationship given a bounding box from a subject $S$ and object $O$, without looking at the image, \ie $\p(P | O, S, I) = \p(P | O, S)$. For the FREQ baseline it is computed using all pairs of boxes in the train set. FREQ-OVERLAP instead is computed using only overlapping pairs of boxes. 
Further, assuming the presence of $O$ is independent of $S$, $\p(O | S, I) = \p(O | I)$.

To compute the $\p(O | I)$ and $\p(S | I)$ factors we use the same pretrained base detection model used to initialize \method (Section~\ref{subsec:impl_details}).
After the set of detection is produced, we derive the final score for each pair of detections according to Eq.~\eqref{eq:baselines} and using the prior (FREQ baseline). For the FREQ-OVERLAP baseline only overlapping pairs of boxes are scored using the corresponding prior. 

\noindent\textbf{Quantitative results.}
We compare the results of our \method, the baselines, and the participants in the \OID VRD Challenge 2018 in Table~\ref{table:res_oid}.
We use values of the public leaderboard on the Kaggle server for validation and report the score both on the public and private leaderboard for all methods. 

When comparing our method to the stronger FREQ-OVERLAP baseline it is clear that using our attention mechanism brings a strong benefit ($+8.1$ score points on the private leaderboard, which corresponds to $+47\%$ relative improvement).
We also compare our approach to the top three challenge participants who used similar architectures as ours for the base detector model (ResNet50+FPN). Our approach outperforms their results as well.

When considering all participants, including those based on more powerful architectures, then our \method is the second best on the current leaderboard according to the private score.
Note, however, that all winning models in the challenge use an ensemble of $2$ models treating attributes prediction separately, while our approach treats both two object relationships and attributes uniformly.
Finally, the number 1 winning model achieves $28.5$ score on the private leaderboard. However, they use much more powerful backbone (ResNet101Xt~\cite{Xie2017cvpr}) and possibly Visual Genome dataset for pretraining their model, so the results are not directly comparable.

%% file: conclusion.tex
\section{Conclusion}

We presented a new model, \method, for detecting visual relationships that relies on a box attention mechanism.
Our model has several important benefits over previously proposed models.
First, it is conceptually simple and theoretically sound: we tackle visual relationship detection by 
formulating it as a task of learning a probabilistic model and
then decomposing this model into simpler sub-models using the chain rule.
Second, our model does not introduce any new hyperparameters on top of those already required by the base detection model it builds on.
Finally, \method delivers strong performance on three challenging datasets.

Overall, our model establishes a strong and easy-to-implement reference for future research. 
Moreover, \method can be an appealing choice 
for practitioners, who want to seamlessly deploy real-life systems for detecting
visual relationships.

\small{\textbf{Acknowledgments.}
We would like to thank Rodrigo Benenson and Jasper Uijlings 
for the helpful feedback during the initial stages of this project.
We also thank Georgia Gkioxari for clarifying our questions
regarding the evaluation protocol.
This work was partially funded by the European Research Council
under the European Unions Seventh Framework Programme (FP7/2007-
2013)/ERC grant agreement no 308036.}